\documentclass[nohyperref]{article}
\usepackage[utf8]{inputenc}
\usepackage{graphicx}
\usepackage{subfigure}
\usepackage{booktabs} 
\usepackage{microtype}
\usepackage{numprint}
\usepackage{textcomp}
\usepackage{hyperref}

\usepackage{tikz-cd}
\usepackage{adjustbox}
\usepackage{dirtytalk}
\usepackage{pifont}
\usepackage{multirow}

\usepackage[accepted]{icml2022}

\usepackage{amsmath}
\usepackage{amssymb}
\usepackage{mathtools}
\usepackage{amsthm}
\usepackage{graphics}
\hypersetup{draft}
\usepackage[capitalize,noabbrev]{cleveref}

\theoremstyle{plain}

\theoremstyle{definition}

\theoremstyle{remark}

\usepackage[textsize=tiny]{todonotes}
\usepackage[acronym,toc,shortcuts]{glossaries}
\newacronym{DP}{DP}{Differential Privacy}
\newacronym{DPSGD}{DP-SGD}{Differentially Private Stochastic Gradient Descent}
\newacronym{BN}{BN}{Batch Normalisation}
\newacronym{IN}{IN}{Instance Normalisation}
\newacronym{LN}{LN}{Layer Normalisation}
\newacronym{GN}{GN}{Group Normalisation}

\icmltitlerunning{Differentially private training of residual networks with scale normalisation}

\begin{document}

\twocolumn[
\icmltitle{Differentially private training of residual networks with scale normalisation}



\icmlsetsymbol{equal}{*}

\begin{icmlauthorlist}
\icmlauthor{Helena Klause}{tum}
\icmlauthor{Alexander Ziller}{tum,rad}
\icmlauthor{Daniel Rueckert}{tum,icl}
\icmlauthor{Kerstin Hammernik}{tum,icl}
\icmlauthor{Georgios Kaissis}{tum,rad,icl}
\end{icmlauthorlist}

\icmlaffiliation{tum}{Chair for AI in Medicine, Technical University of Munich, Germany}
\icmlaffiliation{rad}{Department of Radiology, Technical University of Munich, Germany}
\icmlaffiliation{icl}{Department of Computing, Imperial College London, United Kingdom}

\icmlcorrespondingauthor{Georgios Kaissis}{g.kaissis@tum.de}

\icmlkeywords{Differential Privacy, Deep Learning, Architecture Design}

\vskip 0.3in
]



\printAffiliationsAndNotice{}  

\begin{abstract}
The training of neural networks with Differentially Private Stochastic Gradient Descent offers formal Differential Privacy guarantees but introduces accuracy trade-offs. In this work, we propose to alleviate these trade-offs in residual networks with Group Normalisation through a simple architectural modification termed \textit{ScaleNorm} by which an additional normalisation layer is introduced after the residual block's addition operation. Our method allows us to further improve on the recently reported state-of-the art on CIFAR-10, achieving a top-1 accuracy of 82.5\% ($\varepsilon=8.0$) when trained \textit{from scratch}. 
\end{abstract}

\section{Introduction}
\acrfull*{DPSGD} \citep{abadi2016deep} is arguably the most widely utilised technique for training deep neural networks with \acrfull{DP} guarantees. Although it offers formal guarantees of privacy to the individuals whose data is used to train the model, its utilisation results in accuracy trade-offs. These trade-offs are undesirable, especially in critical domains such as medicine where both high diagnostic accuracy and strong privacy guarantees are required, prompting the development of  methods to alleviate these privacy-utility trade-offs. In the current work, we focus on computer vision tasks, which represent one of the broadest application areas for deep learning. 

Residual networks (ResNets) \citep{he2016deep} are a standard architecture in computer vision. The residual block, their core architectural building block, has since become a standard component of architectures responsible for several victories in the ImageNet (ILSVR) challenge \citep{deng2009imagenet}. Moreover, despite newer works introducing novel architectures which outperform the original ResNets, it has recently been shown that their performance can be comparable to current-generation vision models in challenging tasks \citep{wightman2021resnet} if trained appropriately. The combination of ResNets and \gls*{DPSGD} thus appears as a promising direction for privately training computer vision models. Indeed, both of the very recently reported successful applications of \gls*{DPSGD} to the ImageNet dataset were achieved using some variation of the ResNet architecture \citep{kurakin2022training, de2022_unlocking}.

In the current work, we observe and address a phenomenon arising in the training of ResNets with \gls*{DPSGD}, which we term \textit{scale mixing}: Inside a residual block, the activations flowing through the convolutional path are normalised, however the activations flowing through the residual path are not. This leads to impaired convergence and reduced test-set accuracy both in non-DP training and with DP-SGD, but is substantially more pronounced with DP-SGD. 

\paragraph*{Contributions and overview of the paper} Our key contribution is the introduction of an architectural adaptation to the residual block, which we term \textit{scale normalisation} (ScaleNorm). ScaleNorm fixes the scale mixing problem and empirically leads to improved convergence, which we investigate in Section \ref{sec:scalenorm}. Experimentally, we demonstrate the superiority of ResNets with ScaleNorm to regular ResNets on several image benchmark tasks (CIFAR-10, ImageNette, TinyImageNet) in Section \ref{sec:experiments}, where we show that --combined with recently introduced training adaptations-- we achieve state-of-the-art top-1 accuracy on CIFAR-10 when trained \textit{from scratch}.

\section{Prior work}
The attempt to alleviate privacy-utility trade-offs in \gls*{DPSGD} has been the subject of two distinct lines of work. We will focus exclusively on the works investigating the \gls*{DPSGD} algorithm as presented in \citet{abadi2016deep} and omit works relying on modifications to the \gls*{DPSGD} algorithm or alternative techniques for obtaining improved accuracy (such as \citet{zhu2020private} or \citet{papernot2018scalable}). 

The first approach to improving the training outcome of \gls*{DPSGD} models is \textbf{transfer learning}, that is, \textit{fine-tuning} a network which has been pre-trained on public data \citep{luo2021scalable, davody2021effect, tramer2021differentially}. Such works are able to leverage pre-trained representations (or, in the case of \citet{tramer2021differentially}, fixed features from wavelet transforms). This approach has several major benefits: The quality of non-privately learned representations is higher, enabling the practitioner to only train a subset of the network's layers with DP-SGD, which is both faster and leads to improved accuracy. Moreover, Batch Normalisation (BN) layers can be used in the part of the network that is not trained, as they remain \textit{frozen}.

Our work studies a different situation in which transfer learning is not possible or ineffective, and networks are trained \textit{from scratch}. Such a scenario can arise in cases of a large domain discrepancy between the pre-training dataset and the task at hand (e.g. high-dimensional medical imaging) or because the pre-training dataset is itself sensitive in nature. Moreover, omitting pre-training allows us to empirically observe the training behaviour of the network, which is of independent interest. Two possibilities exist to improve the convergence properties and test-set accuracy. The first are \textbf{training adaptations} such as large-batch training \citep{doermann2021noise}, augmentation multiplicity \citep{fort2021_drawing} or Polyak-Ruppert/exponential weight averaging \citep{polyak1992acceleration, ruppert1988efficient, tan2019efficientnet}. These techniques were very recently combined by \cite{de2022_unlocking} to achieve the state-of-the art with DP-SGD on ImageNet and CIFAR-10. The second possibility are \textbf{architectural adaptations}. These can take the form of adapting the architecture from the ground up, such as \citet{morsbach2021architecture} and \citet{papernot2020making}, who support the hypothesis that specialised architectures for use with DP are required to be designed. A less complicated route is making modifications to \textit{existing} architectures, for example by replacing Batch Normalisation with alternatives such as Group Normalisation (which is mandatory for maintaining DP guarantees) or exchanging the activation functions \citep{papernot2020tempered}. Our work takes cues from both aforementioned categories, as we modify an existing architecture to improve its performance under DP. In addition, we combine the resulting improved architecture with the training adaptations presented by \citet{de2022_unlocking} to further improve our results.

\section{Methods and Results}

\subsection{Scale mixing and ScaleNorm}
\label{sec:scalenorm}
\begin{figure}[h]
    \centering
    \includegraphics[scale=0.08]{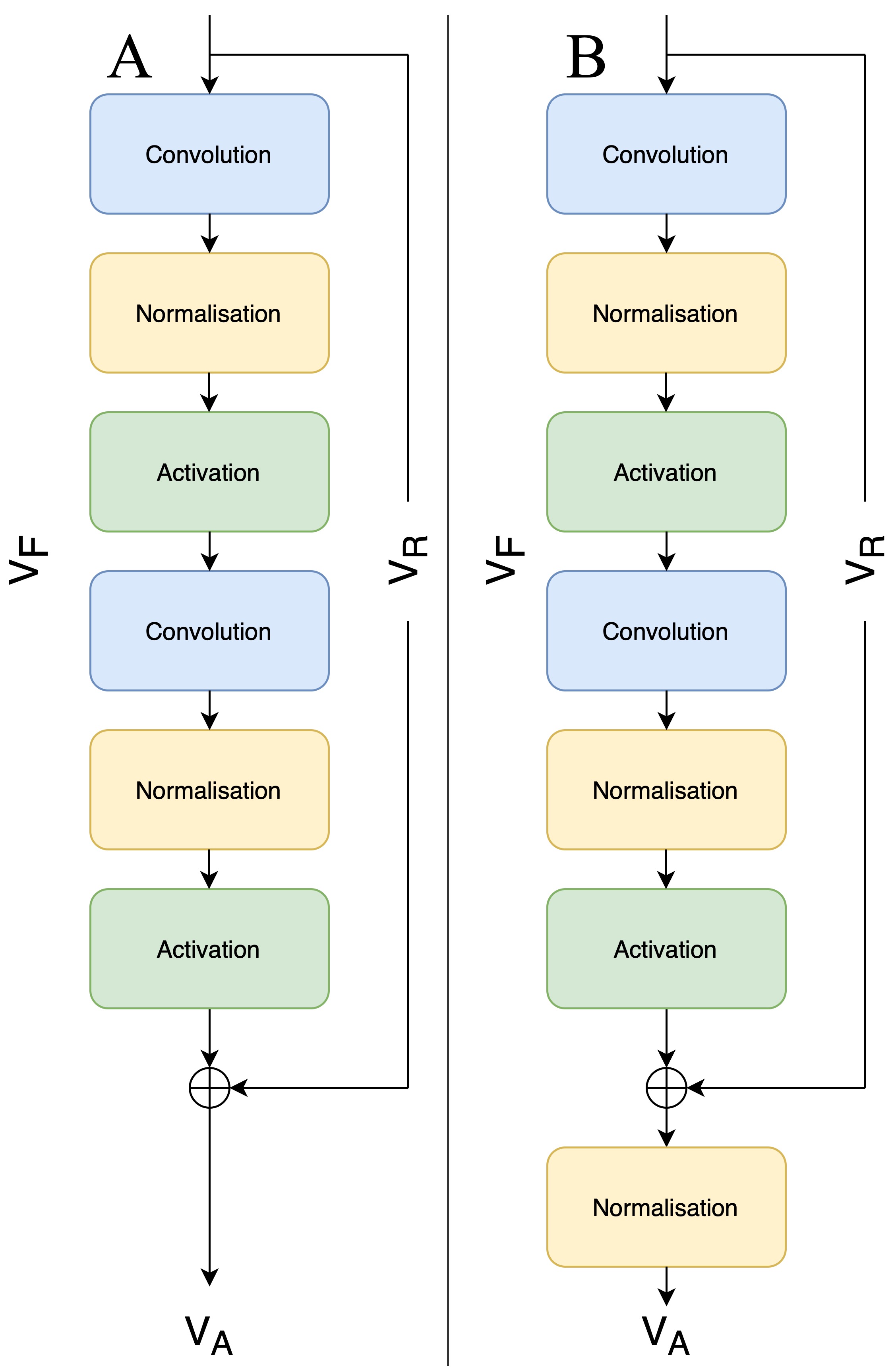}
    \caption{Schematic representation of a residual block (\textbf{A}) vs. a residual block with ScaleNorm (\textbf{B}). Observe the additional normalisation after the addition operation. $V_R$: Residual path, $V_F$: Convolutional path, $V_A$: Activation.}
    \label{fig:resblock}
\end{figure}

\begin{figure*}[h]
\centering
\includegraphics[width=0.85\textwidth]{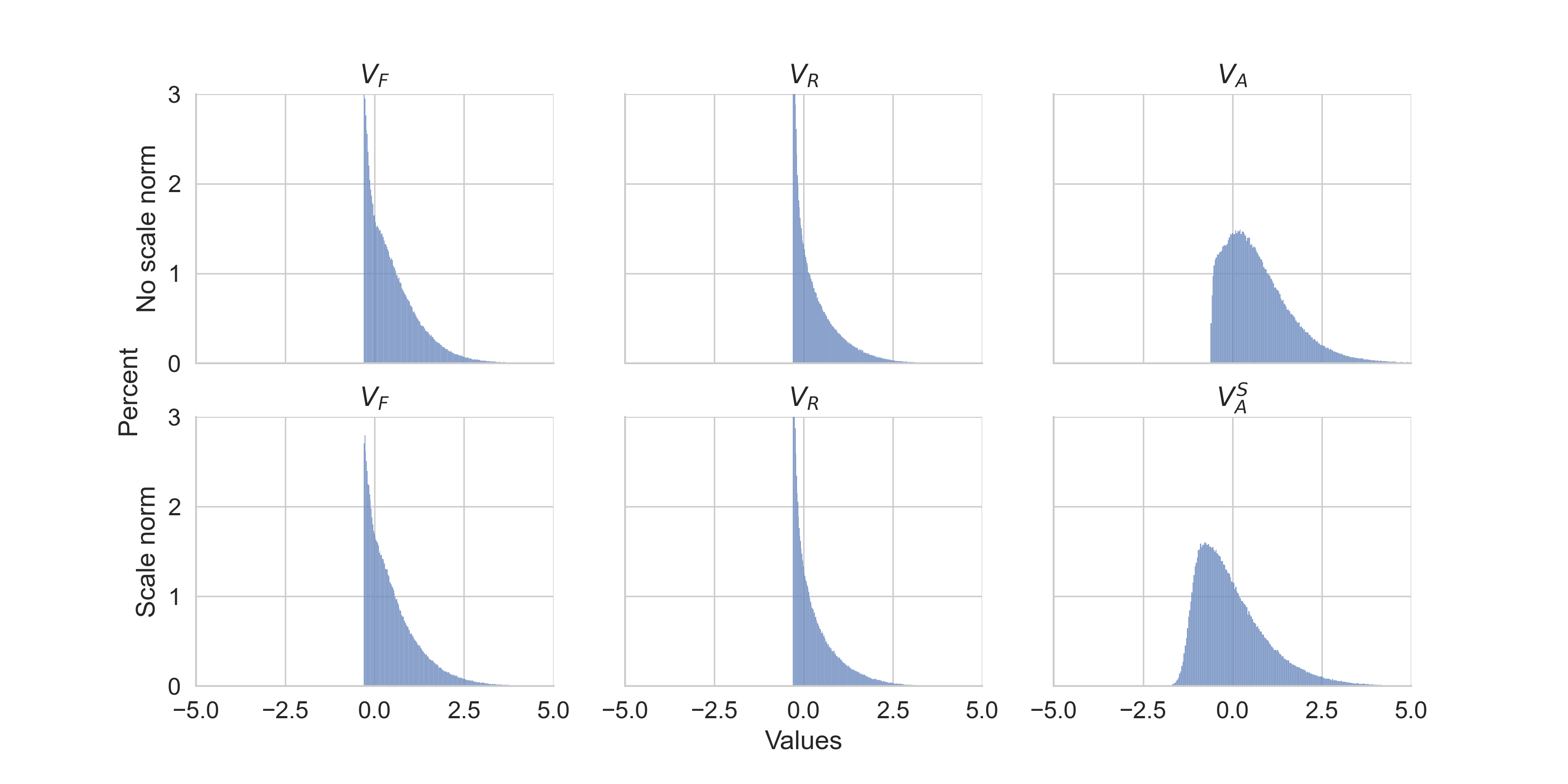}
\caption{Activation histograms of a ResNet (top row) vs. a ScaleResNet (bottom row) residual block. Observe that the activations output by $V_A$ (top right) are markedly asymmetric with substantial mass on the positive side (sample average: 0.72, standard deviation: 0.76), whereas the activations output by the scale normalisation operation $V^S_A$ (bottom right) are more symmetric about the sample mean of 0 and have unity standard deviation.}
\label{fig:hists}
\end{figure*}

\begin{figure}[h]
    \centering
    \includegraphics[scale=0.50]{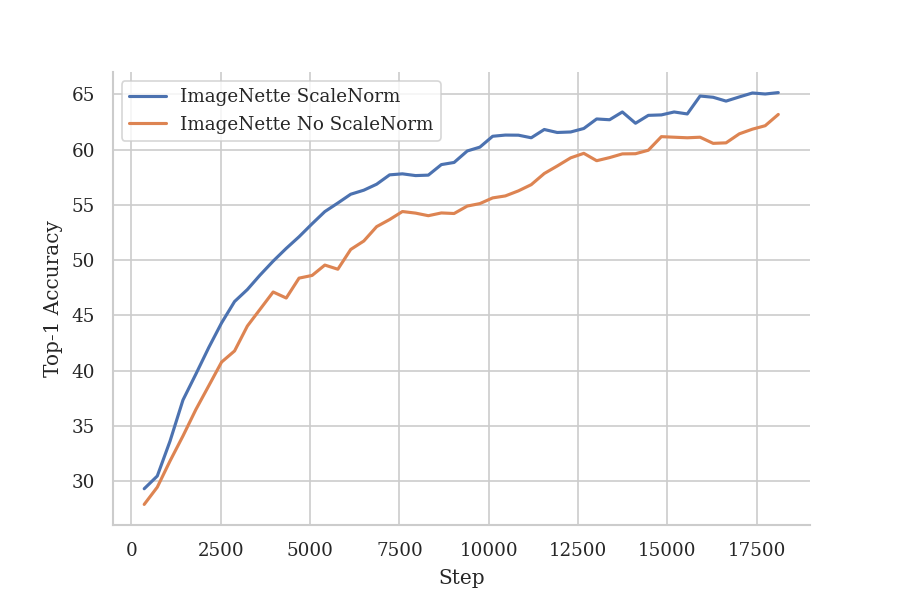}
    \caption{Exemplary training dynamics of a ScaleResNet (blue curve) vs. a regular ResNet (orange curve) in terms of test accuracy on the ImageNette dataset. Observe the markedly improved convergence properties of the ScaleResNet.}
    \label{fig:convergence}
\end{figure}

We begin by discussing the origins of the scale mixing problem in ResNets. Figure \ref{fig:resblock}A shows a conventional residual block. The residual path's activations ($V_R$) are the immediate output of the previous layer and are scaled (i.e. distributed) differently from the convolutional path's activations ($V_F$), which undergo normalisation twice. The addition operation at the end of the block \textit{mixes} the scales of the two activations, which skews the distributions of the outputs ($V_A$). Figure \ref{fig:hists} (top row) demonstrates this behaviour. ScaleNorm modifies the residual block to include an additional normalisation layer after the addition function (Figure \ref{fig:resblock}B), which re-normalises the outgoing activations $V_A$. Empirically, this has several beneficial effects. It renders the activations $V_A$ more symmetric about the mean and returns their standard deviation to 1, which has been noted in previous works as a desirable property for training stability \citep{glorot2010understanding, zhang2018residual}. Moreover, we experimentally found ResNets with ScaleNorm (ScaleResNets) to show substantially improved convergence compared to regular ResNets, as exemplarily shown in Figure \ref{fig:convergence}. This knowledge enables practitioners to \textit{save privacy budget} by training their networks for fewer epochs or with more noise. Beyond improving the activation distributions, we conjecture ScaleNorm to also smoothen the optimisation landscape. Experimentally, the ScaleResNet-9 architecture used in the experiments below has a significantly lower value of the Hessian trace (6637.3) compared to the regular ResNet-9 (9045.7), a slightly lower condition number (\numprint{2.1e-3} vs \numprint{2.3e-3}), fewer negative Hessian eigenvalues (115 vs. 119) and lower maximum Hessian eigenvalues (2654.2 vs. 3920.4), indicating a lower curvature of the loss landscape at the convergence point. Of note, ScaleResNets consistently performed better than regular ResNets even without DP-SGD, but the effect was negligible (ca. 0.5\% accuracy advantage), which explains why this architectural modification is not standard in non-private deep learning literature. Details of the loss landscape experiments and non-private training can be found in the Appendix. 

\subsection{Experiments}
\label{sec:experiments}
We assume familiarity with the DP-SGD algorithm \citep{abadi2016deep}. Unless otherwise indicated, all experiments below were conducted in triplicate with fixed random seeds selected randomly in advance. 

\subsubsection{Experiments with (Scale)ResNet-9}
DP-SGD is resource-intensive and the noise added per step scales proportionally to the number of (trainable) model parameters \citep{subramani2021enabling}. Thus, we chose to construct a shallow nine-layer ResNet with \numprint{2447946} parameters for the following experiments. We term this architecture ResNet-9, or ScaleResNet-9 when ScaleNorm was used. The architecture includes Group Normalisation (GN) layers per default. Details of the architectural design and a comparison with ResNet-50, supporting the choice of a smaller model for the following experimental evaluation, can be found in the Appendix. We evaluated the effect of ScaleNorm on CIFAR-10 at an image size of 32\texttimes32 \citep{krizhevsky2009learning}, Tiny ImageNet \footnote{available from \url{http://cs231n.stanford.edu/}} at a size of 64\texttimes64 and the 2019 version of ImageNette \citep{imagenette}, a subset of 10 ImageNet classes, at an image size of 160\texttimes160. We trained the CIFAR-10 and ImageNette networks for 50 epochs and the Tiny ImageNet network for 90 epochs with an $L_2$-norm bound (clipping threshold) of 1.5 using an expected lot (batch) size of 1024 with the NAdam optimiser \cite{dozat2016nadam} at a learning rate of 0.001. We employed a constant learning rate schedule with reduction by half upon stagnation of the validation loss for more than three epochs. We performed a hyperparameter search over the optimal number of GN groups from 1 (equivalent to Layer Normalisation) to the number of channels (equivalent to Instance Normalisation) and intermediate values of 16, 32, 64. Hyperparameter searches were conducted using the \textit{Optuna} package \cite{optuna_2019} with the \textit{Tree-structured Parzen Estimator} search strategy and the \textit{threshold} pruner. We reset the privacy budget for each hyperparameter search run similar to \citet{kurakin2022training, de2022_unlocking}, which we note to not preserve DP. $\varepsilon$ was calculated at a $\delta$ of \numprint{e-5} using the Rényi DP accountant of the \textit{Opacus} package \citep{opacus}. We performed the experiments on CIFAR-10 and ImageNette at three privacy levels: 2.89 (similar to \citet{tramer2021differentially}), 7.42 (similar to \citet{doermann2021noise}) and 9.88 (similar to \citet{kurakin2022training}). Tiny ImageNet was trained at $\varepsilon \in$ (5, 10, 70) similar to \citet{kurakin2022training} because the dataset is very difficult to train to acceptable accuracy from scratch with DP-SGD with lower $\varepsilon$, similar to ImageNet itself, as seen in \citet{kurakin2022training, de2022_unlocking}. Table \ref{tab:result_summary} summarises our results on CIFAR-10, ImageNette and Tiny ImageNet, respectively.  

\begin{table}[ht]
\centering
\begin{tabular}{@{}llll@{}}
\toprule
\multicolumn{4}{c}{CIFAR-10}                                       \\ \midrule
$\varepsilon$                & 2.89          & 7.42          & 9.88          \\ \cmidrule(l){2-4} 
ScaleNorm          & \textbf{65.6} & \textbf{71.8} & \textbf{73.0} \\
No ScaleNorm       & 65.0          & 71.4          & 72.4          \\ \midrule \addlinespace[0.5em]
\multicolumn{4}{c}{ImageNette}                                     \\ \midrule
$\varepsilon$                & 2.89          & 7.42          & 9.88          \\ \cmidrule(l){2-4} 
ScaleNorm          & \textbf{56.6} & \textbf{64.8} & \textbf{67.1} \\
No ScaleNorm       & 55.5          & 63.8          & 65.0          \\ \midrule \addlinespace[0.5em]
\multicolumn{4}{c}{Tiny ImageNet}                                  \\ \midrule
$\varepsilon$                & 5             & 10            & 70            \\ \cmidrule(l){2-4} 
ScaleNorm Top-1    & \textbf{15.2} & \textbf{19.4} & \textbf{25.8} \\
No ScaleNorm Top-1 & 14.2          & 18.7          & 24.7          \\
ScaleNorm Top-5    & \textbf{36.0} & \textbf{41.8} & \textbf{50.8} \\
No ScaleNorm Top-5 & 34.4          & 40.5          & 48.3          \\ \bottomrule
\end{tabular}
\caption{Summary of results on CIFAR-10, ImageNette and Tiny ImageNet. All results are median accuracies (in \%) of three runs.}
\label{tab:result_summary}
\end{table}

ScaleResNet-9 consistently outperformed the regular ResNet on all tasks. An even stronger separation was observed when comparing the best (instead of the median) models of the three repetitions; these results can be found in the Appendix.

\subsubsection{Experiments with (Scale)WideResNet-16/4}
During the preparation of our manuscript, \citet{de2022_unlocking} published a study utilising a number of training adaptations on top of the WideResNet (WRN) architecture with 16 layers and a width factor of 4 (WRN-16/4, \numprint{2752506} parameters) to obtain a top-1 accuracy of 81.4\% at an $\varepsilon$ of 8.0. Examining Table 2 of \citet{de2022_unlocking}, we observe that the greatest accuracy improvement is derived from applying the \textit{augmentation multiplicity} and \textit{weight averaging} training adaptations, whereas the \say{baseline} accuracy without these adaptations is comparable to our results mentioned above (71.2\%). To therefore examine whether larger ResNets also benefit from ScaleNorm and from incorporating the aforementioned training adaptations, we utilised the identical WRN-16/4 architecture, which we also modified to include ScaleNorm after each residual block (ScaleWRN-16/14), and identical training settings. These results were tested on the CIFAR-10 test set and are summarised in Table \ref{tab:wrn}.

\begin{table}[h]
\centering
\begin{tabular}{@{}ccc@{}}
\toprule
              & \multicolumn{1}{c}{\begin{tabular}[c]{@{}c@{}}WRN-16/4\\ ($\varepsilon$= 8.0)\end{tabular}} & \multicolumn{1}{c}{\begin{tabular}[c]{@{}c@{}}ResNet-9\\ ($\varepsilon$= 7.42)\end{tabular}} \\ \midrule
ScaleNorm     & \textbf{82.5}& \textbf{71.8}\\
No ScaleNorm  & 81.25& 71.4\\
Training time & 963 min& 22 min\\ \bottomrule
\end{tabular}
\caption{Accuracy (in \%) and training time comparison between ResNet-9 and WideResNet-16/4 with and without ScaleNorm. WRN-16/4 was trained with the adaptations proposed by \citet{de2022_unlocking}, which drastically increase training time.}
\label{tab:wrn}
\end{table}

The ScaleWRN-16/4 architecture achieved an accuracy of 82.5\%, outperforming the standard WRN-16/4 (81.25\%) and further improving on the results by \citet{de2022_unlocking} to achieve a (to our knowledge) new state of the art result. We remark that training the ScaleResNet-9 architecture discussed in the previous section requires approximately 20 minutes on a single NVidia Quadro RTX 8000 GPU. Training with the adaptations proposed by \citet{de2022_unlocking} incurs a massive increase in training time to over 16 hours on a dual-GPU system, which should be weighed against the ca. 10\% accuracy increase from a resource efficiency point of view, and prompts further investigation into software and hardware optimisations for DP-SGD.

\section{Discussion and Conclusion}
The broad implementation of \gls*{DP} to large-scale computer vision tasks will require tackling the privacy-utility trade-offs inherent to \gls*{DPSGD}. Our work introduces ScaleNorm, an architectural modification which substantially improves the accuracy of ResNets trained with \gls*{DP}. In future work, we intend to supplement our findings with data on similar architectures such as DenseNets, to clarify the interplay of ScaleNorm with other architectural components such as activation functions or initialisation, and to theoretically investigate its effect on training dynamics.

\bibliography{example_paper}
\bibliographystyle{icml2022}

\newpage
\appendix
\onecolumn
\section*{Appendix}
\begin{small}
\section{Details on training dynamics and Hessian analysis}
To analyse the Hessian of the trained network, we used the \textit{PyHessian} package \citep{Yao2019-mc}. For eigenvalue computation, we used the power iteration method with 1000 iterations and a maximum tolerance of 0.001. For computing the Hessian trace, we used Hutchinson's method with 1000 iterations and a tolerance of 0.001.

\section{Non-private ScaleResNet training}
To assess whether ScaleNorm also improves convergence and test-set accuracy in the non-private setting, we trained the ResNet-9 architecture with Batch Normalisation (instead of the default Group Normalisation) on the CIFAR-10 dataset. Over three runs, the ScaleResNet-9 achieved a median test-set accuracy of 89.8, whereas the regular ResNet-9 reached a test set accuracy of 89.3. We empirically observed the training of the ScaleResNet to be slightly more stable, which we attributed to the additional mild regularising effect of the additional BN layers. Through this experiment, we empirically conclude that --whereas ScaleNorm minimally benefits non-DP training-- further evaluation is required to assess whether a recommendation to use ScaleNorm outside of DP-SGD is warranted.

\section{Details on the ResNet-9 architecture}
The ResNet-9 architecture consists of nine-layers and includes \numprint{2447946} parameters. The network consists of two convolutional blocks (64 and 128 filters) followed by one residual block with 128 filters, two convolutional blocks with 256 filters each, a residual block with 256 filters, a global Max Pooling layer and a fully connected layer with 1024 units. We utilised \textit{Mish} activation functions \citep{misra2020mish} throughout the network. Every convolutional block consists of a convolutional layer followed by an activation and a Group Normalisation layer. For networks with ScaleNorm, the additional normalisation operations follow directly after the residual blocks. An implementation of the architecture using the PyTorch machine learning package can be found at \url{https://gist.github.com/gkaissis/6db6b7271f93d3459263b6978cfd4146}. We note that the code assumes a CPython version $\geq$3.9. Users of prior versions must replace the type annotation on line 41 with \texttt{typing.Tuple}.

\section{Comparison with ResNet-50}
We experimentally compared the performance of ResNet-9 to ResNet-50 on the ImageNette dataset to investigate whether the larger architecture would offer benefits due to higher learning capacity. We found that, although ScaleResNet-50 outperformed the regular ResNet-50, the drop in performance due to the high parameterisation and resulting high noise levels coupled with a much slower training does not justify its use (Table \ref{tab:resnet50}).

\begin{table}[h]
\centering
\begin{tabular}{@{}ll@{}}
\toprule
             & ImageNette \\ \midrule
ScaleNorm    & \textbf{44.2}       \\
No ScaleNorm & 43.0       \\ \bottomrule
\end{tabular}
\caption{Results of ResNet-50 and ScaleResNet-50 on ImageNette. $\varepsilon$=7.42, 50 epochs, median of three runs.}
\label{tab:resnet50}
\end{table}

\section{Best-of-three}
For completeness, we include the results of the best-performing ScaleResNet-9s, which typically showed larger separation from their regular counterparts compared to the median shown in the main manuscript, as seen exemplarily in Table \ref{tab:max_perf}.

\begin{table}[h]
\centering
\begin{tabular}{@{}lccc@{}}
\toprule
             & \multicolumn{1}{c}{\begin{tabular}[c]{@{}c@{}}CIFAR-10 \\ ($\varepsilon$ 7.42)\end{tabular}} & \multicolumn{1}{c}{\begin{tabular}[c]{@{}c@{}}ImageNette\\ ($\varepsilon$ 7.42)\end{tabular}} & \multicolumn{1}{c}{\begin{tabular}[c]{@{}c@{}}Tiny ImageNet\\ ($\varepsilon$ 10)\end{tabular}} \\ \midrule
ScaleNorm    & \textbf{72.4}& \textbf{66.1}& \textbf{42.8}\\
No ScaleNorm & 71.5& 64.4& 41.2\\ \bottomrule
\end{tabular}
\caption{\textit{Best of three runs} classification performance on CIFAR-10, ImageNette and Tiny ImageNet (top-5).}
\label{tab:max_perf}
\end{table}
\end{small}


\end{document}